\documentclass[12pt]{article}

\usepackage[utf8]{inputenc}
\usepackage[T1]{fontenc}
\usepackage{amsmath,amssymb}
\usepackage{graphicx}
\usepackage{booktabs}
\usepackage{array}
\usepackage{longtable}
\usepackage{hyperref}
\usepackage{url}
\usepackage{geometry}
\usepackage{multirow}
\usepackage{authblk}
\usepackage{microtype}
\usepackage{parskip}
\usepackage{listings}
\usepackage{xcolor}
\usepackage{float}
\usepackage{caption}
\usepackage{subcaption}
\usepackage{natbib}
\usepackage{tikz}
\usetikzlibrary{arrows.meta,positioning,shapes.geometric}

\title{\textbf{Solar Intelligence}}

\author[1]{Jyotsna Singh\thanks{\href{mailto:jsinghenv@gmail.com}{jsinghenv@gmail.com}}}

\affil[1]{College of Information Science, University of Arizona, Tucson, AZ, USA}

\date{}

\usepackage{xcolor}
\usepackage{tcolorbox}
\tcbuselibrary{skins, breakable}

\newtcolorbox{querybox}[1][]{
  colback=gray!8,
  colframe=gray!45,
  boxrule=0.4pt,
  arc=2pt,
  left=10pt, right=10pt,
  top=8pt, bottom=8pt,
  fontupper=\small,
  breakable,
  title=#1,
  fonttitle=\bfseries\small,
  coltitle=black,
  colbacktitle=gray!25
}
\begin{document}

\maketitle

\begin{abstract}
Solar energy decision support is fragmented across dashboards that provide data without explanation, research papers are slow to parse, and general-purpose language models are not solar domain specific and answer without evidence. This paper introduces Solar Intelligence, a hybrid retrieval-augmented framework that unifies structured solar analytics, evidence-grounded scientific question answering, and machine learning forecasting in one system. The platform integrates daily NASA POWER solar and meteorological data, Biosphere 2 ground-sensor readings, and a curated corpus of research papers and institutional reports. Structured queries use DuckDB SQL; scientific questions are answered by a hybrid retriever that fuses BM25 and ChromaDB dense embeddings via Reciprocal Rank Fusion, with responses grounded through a language model (llama3.2:3b). An Extreme Gradient Boosting (XGBoost) model produces daily forecasts of irradiance, temperature, and wind speed. The system is exposed via FastAPI, Streamlit, and an MCP server, so it can be used as an application, an API service, or an agent tool - by students, researchers, and energy analysts.
\end{abstract}


\section{Introduction}

\qquad A reliable decision in the field of solar energy needs domain knowledge, the latest research, available data to understand the status, and a forecast to understand the future. This complete combination allows a user to evaluate holistically, so that solar energy systems can be planned, organized, and maintained. Solar energy system installation planning depends on long-term irradiance variability, seasonal trends, and unusual pattern detection (Segado-Moreno et al., 2026; Singh and Kumar, 2016; Wild et al., 2009). Scientists and researchers have built models ranging from empirical to Machine Learning (ML) to predict global (Fohagui et al., 2026; Voyant et al., 2014) and diffuse solar radiation (Appelbaum and Peled, 2026; Atsbeha et al., 2025; Singh et al., 2013). Data from photovoltaic (PV) systems is also being analysed to understand solar energy generation from these systems (Ouyang et al., 2025; Singh et al., 2017).

\qquad Many data dashboards, forecasting services, and Large Language Models (LLMs) are already helping in solar radiation research, but they are fragmented, and this fragmentation delays both research and decision making. Gathering knowledge from research papers and reports takes a lot of time, and traditional statistical and ML models can analyse historical solar radiation patterns but do not support natural-language querying or provide evidence-grounded explanations. When an engineer, a scientist, or a student starts to work in this domain, they have to move between different dashboards and forecasting websites, request data from data companies, analyse it, and only then take a decision--steps that involve months to a year of time. Recent advances in Retrieval-Augmented Generation (RAG) demonstrate that combining information retrieval with LLMs enables knowledge-intensive reasoning grounded in external data sources (Lewis et al., 2020), and recent surveys highlight that RAG frameworks address the limitations of conventional models by reducing hallucination and improving factual consistency through explicit retrieval mechanisms (Gao et al., 2023).

\qquad This project proposes Solar Intelligence; an intelligent system built on a Hybrid RAG framework. The system integrates multi-year solar irradiance datasets with curated scientific literature to enable both structured and semantic retrieval. It retrieves relevant data windows and scientific document passages, provides them as grounded context to an LLM, and generates cited, evidence-based responses through an interactive Streamlit dashboard. This is important because modern solar operations depend on forecast quality , not just data access, and recent energy forecasting studies show that forecast quality directly affects grid operations, scheduling, and economic value in solar power systems (Mayer et al., 2026; Gandhi et al., 2024).

\qquad The main goal of Solar Intelligence is to build one software platform for solar domain querying, analytics, and forecasting, so that a user does not have to move between separate solar radiation tools to reach a decision. The system was designed with three objectives. The first is to combine multi-year solar data and a curated body of scientific literature in a single retrieval system, so that numerical evidence and published evidence are reachable through the same query. The second is to support natural-language questions on solar energy topics and to return answers that are grounded in the retrieved data and documents and carry citations, which reduces unsupported responses and removes the need for the user to know the structure of the underlying data. The third is to provide forecasting for important solar energy variables within the same interface used for querying. The system is delivered as working software, accessible through both a dashboard and an API, and built on a modular design that can be updated and scaled. Building the retrieval layer on a domain-specific corpus of solar radiation literature, rather than on general web knowledge, distinguishes this system from a general-purpose language model.

\qquad The present work is organized as follows. After the introduction, the datasets used in this study are described, followed by the methodology used to build the system and then the system architecture and implementation. The outputs of the Solar Intelligence query system and forecasting are presented in the results and discussion section, which is followed by the deployment architecture, the limitations of the current version of Solar Intelligence, the conclusions, and the future work.

\section{Data Sources}

\qquad Solar Intelligence integrates three complementary data sources (Table~\ref{tab:data}). The first is a domain-specific corpus of 149 research papers and reports (1994--2026) covering solar energy science, engineering, and instrumentation, comprising 138 arXiv papers retrieved across 18 topic-based queries and 11 institutional reports from NREL, IEA-PVPS, IRENA, and Sandia. The second is daily meteorological and solar radiation data (2012--2026, 5,225 daily records) obtained from the NASA POWER data service for Biosphere 2, Oracle, Arizona (32.578$^{\circ}$N, 110.848$^{\circ}$W) location, providing solar irradiance, air temperature, relative humidity, wind speed, and precipitation. Within NASA POWER, solar radiation is derived from the CERES SYN1deg satellite product, and meteorological parameters are derived from the Modern-Era Retrospective analysis for Research and Applications, Version 2 (MERRA-2) reanalysis produced by NASA's Global Modeling and Assimilation Office. The third is ground-sensor data from the University of Arizona Biosphere 2 facility (August 2012 -- July 2013, 10.4 million minute-resolution readings) covering air temperature, solar irradiance, and wind speed. Here, B2 and b2 both refer to Biosphere~2.

\begin{table}[H]
\centering
\caption{Data sources used in Solar Intelligence}
\label{tab:data}
\begin{tabular}{clll}
\toprule
\textbf{S. No.} & \textbf{Data Details} & \textbf{Data Period} & \textbf{Data Sources} \\
\midrule
1 & Research papers and Reports & 1994--2006 & arXiv, NREL, IEA-PVPS, \\
  &                             &            & IRENA, and Sandia \\
2 & Irradiance, Temperature, Humidity, & 2012--2026 & NASA POWER \\
  & Wind Speed, Precipitation          &            & \\
3 & Air Temperature, Irradiance, and & Aug 2012 -- & University of Arizona \\
  & Wind Speed                       & July 2013   & Biosphere 2 \\
\bottomrule
\end{tabular}
\end{table}

\section{Methodology}

\qquad The methodology consists of five components: corpus construction and processing, retrieval through query routing, Biosphere 2 ground-sensor data cleaning, cross-source comparison between the satellite product and the ground sensors and, forecasting. Each component is described in the subsections below.

\subsection{Corpus Construction and Processing}

\qquad The domain specific (solar energy) corpus was constructed with research papers and reports. The arXiv subset is built by issuing topic-based queries via the arXiv API across 18 different solar subdomains, including photovoltaic efficiency, irradiance forecasting, soiling effects, sky imager-based prediction, atmospheric attenuation, and concentrator photovoltaics (Table~\ref{tab:queries}). The institutional subset is downloaded directly from NREL, IEA-PVPS, IRENA, and Sandia, including the System Advisor Model reference manual and the PVWatts version 5 manual.

\begin{table}[H]
\centering
\caption{Eighteen topic queries used to construct the arXiv subset of the corpus via the official arXiv API}
\label{tab:queries}
\begin{tabular}{c p{3.3cm} p{6.2cm} c}
\toprule
& \textbf{Category} & \textbf{Topic Query} & \textbf{Max Results} \\
\midrule
1 & Forecasting \& ML & solar irradiance forecasting machine learning & 20 \\
  &                   & solar power generation neural network & 15 \\
\addlinespace
2 & Irradiance components & global horizontal irradiance GHI estimation & 15 \\
  &                       & direct normal irradiance DNI clear sky & 10 \\
  &                       & solar radiation satellite measurement validation & 15 \\
  &                       & solar resource assessment NASA POWER & 10 \\
\addlinespace
3 & PV physics & photovoltaic efficiency temperature coefficient & 15 \\
  &            & solar cell efficiency limit Shockley-Queisser & 10 \\
  &            & photovoltaic system performance modeling & 15 \\
\addlinespace
4 & Degradation \& losses & PV soiling dust desert climate & 10 \\
  &                       & photovoltaic degradation long-term & 10 \\
  &                       & solar inverter MPPT efficiency & 10 \\
\addlinespace
6 & Atmospheric / meteorological & atmospheric aerosol solar radiation & 10 \\
  &                              & solar energy meteorology cloud cover & 10 \\
\addlinespace
7 & Advanced PV configurations & bifacial solar panel albedo & 10 \\
  &                            & concentrated solar power CSP thermal & 10 \\
\addlinespace
8 & Standards \& instrumentation & solar spectrum AM1.5 standard & 8 \\
  &                              & pyranometer ground measurement validation & 10 \\
\midrule
\multicolumn{3}{r}{\textbf{Total requested}} & \textbf{213} \\
\multicolumn{3}{r}{\textbf{After deduplication}} & \textbf{138} \\
\bottomrule
\end{tabular}
\end{table}

\qquad PDFs are extracted using PyMuPDF and split into approximately 500-token chunks with 50-token overlap (9,496 chunks total). Two parallel indices are built: 1) BM25 (Robertson and Walker, 1994; Robertson et al., 1994), Tokenization uses lowercase normalization; stemming is deliberately omitted to protect domain-specific tokens (GHI, DNI, MPPT, AM1.5, Shockley-Queisser, perovskite-silicon). 2) Dense embeddings using BGE (small) 384 dimensional (Xiao et al., 2023).

\subsection{Retrieval through query routing}

\qquad Each query is dispatched in parallel to both indices, each returning top-20 candidates. Because BM25 scores and cosine similarities ($[-1, 1]$) are not directly comparable, results are fused via Reciprocal Rank Fusion (Cormack et al, 2009). Not all queries should hit retrieval. A rule-based classifier inspects the query for the presence of metrics (e.g., ``irradiance'', ``temperature''), aggregation keywords (avg/max/min/trend), date ranges, and forecast keywords (forecast/predict/next-day), then assigns one of the following routing tags:

\begin{itemize}
  \item \textit{structured}--pure numeric query, route to DuckDB SQL only.
  \item \textit{semantic}--pure conceptual query, route to hybrid retrieval only.
  \item \textit{hybrid}--combined query requiring both numerical and conceptual evidence; both paths execute.
\end{itemize}

\qquad The classifier also performs source detection (NASA, Biosphere 2 and comparison) based on lexical cues such as ``satellite'', ``ground sensor'', or year ranges (2012--2013 dates auto-suggest B2 comparison). This routing layer is essential because feeding a numerical aggregate question through a generative LLM and sending a conceptual question through SQL is error prone. The natural language answers are composed by llama3.2:3b served via Ollama and follows prompt based strict rules like never invent numerical values and cite every claim with [N] markers. The safety for the system also applied because a small LLM may still hedge exact SQL values, the pipeline orchestrator constructs numerical answer strings deterministically from SQL output and bypasses the LLM entirely for the numeric portion when the classifier flags a strong structured signal. The LLM contributes only prose framing. This makes hallucination prevention architectural rather than purely prompt based.

\subsection{Biosphere 2 Data Cleaning}

\qquad The Biosphere 2 raw data contained implausible values for several meteorological and physical parameters, inconsistent with any real-world atmospheric or electrical phenomenon. Possible causes include sensor faults, communication errors, and unit-conversion artifacts. The selected meteorological variables were therefore range-filtered to maintain physical consistency.
\subsection{NASA POWER and Biosphere 2 Data Comparison}

\qquad The NASA POWER satellite values compared against cleaned Biosphere 2 ground sensor values across 312 overlapping days (August 2012 -- July 2013) for three meteorological variables: irradiance (kWh\,m$^{-2}$\,day$^{-1}$), temperature ($^{\circ}$C), and wind speed (m\,s$^{-1}$). Metrics computed: mean of each source, signed bias (NASA mean $-$ B2 mean) and Mean Absolute Error.
\subsection{Forecasting}

\qquad Daily forecasts for irradiance, temperature, and wind speed are produced by Extreme Gradient Boosting (XGBoost) (Chen and Guestrin, 2016), selected over deep-learning alternatives because tree-based methods perform well on small-to-medium tabular time series (Grinsztajn et al., 2022). The forecaster is trained only on NASA POWER data; the 11-month Biosphere 2 record is insufficient for daily-resolution forecasting requiring multi-year seasonal cycles. Features and hyperparameters of Solar Intelligence are given in Table~\ref{tab:xgb}.

\begin{table}[H]
\centering
\caption{Features used in forecasting}
\label{tab:xgb}
\begin{tabular}{c p{5.5cm} p{4.0cm}}
\toprule
\textbf{S. No.} & \textbf{Features} & \textbf{Types} \\
\midrule
1  & Day-of-year (1--366)                    & \multirow{2}{*}{Calendar features} \\
2  & Month (1--12)                           & \\
\midrule
3  & $\sin(2\pi \cdot \mathrm{doy}/365.25)$  & \multirow{2}{*}{Cyclic seasonality} \\
4  & $\cos(2\pi \cdot \mathrm{doy}/365.25)$  & \\
\midrule
5  & Target value at lag 1 day               & \multirow{4}{*}{Lag features} \\
6  & Target value at lag 7 days              & \\
7  & Target value at lag 30 days             & \\
8  & Target value at lag 365 days            & \\
\midrule
9  & 7-day trailing mean                     & \multirow{2}{*}{Rolling means} \\
10 & 30-day trailing mean                    & \\
\bottomrule
\end{tabular}
\end{table}
\section{System Architecture and Implementation}

\qquad Solar Intelligence is built using modular software architecture. It has four main layers: data layer, analytics and knowledge retrieval layer, forecasting layer, and interface layer. The system configuration and technology stack are summarized in Table~\ref{tab:stack}.

\subsection{Data Layer}

\qquad This layer stores both structured solar data (NASA POWER and Biosphere 2) and unstructured scientific text. The structured data was stored in DuckDB and it includes solar irradiance, temperature, humidity, wind speed, and precipitation. The document layer stores scientific text and solar domain documents.

\subsection{Analytics and Knowledge Retrieval Layer}

\qquad This layer handles the query understanding and execution. When a user asks a question, the system first check the type of question and decides whether it is a structured, semantic and hybrid query. This check is important because all the queries are not handled the same way. A numerical question should not go directly to a language model and a scientific question can't be handled only using SQL. Structured queries are routed to DuckDB and answered using SQL. Semantic queries are routed to the retrieval system and answered using documents. Hybrid queries use both SQL and document retrieval together. The retrieval system uses BM25 and semantic embeddings together. Results from both are combined using Reciprocal Rank Fusion (RRF). This improves retrieval quality by combining exact keyword matching and semantic meaning. After retrieval, the grounded generation module builds the final answer. This module does not answer from model memory alone. It uses structured outputs, retrieved evidence and strict prompt rules. This reduces unsupported responses and improves traceability. It also reduces the model hallucination and provide users with reliable outputs and also guide them if they ask a random question.

\subsection{Forecasting Layer}

\qquad The forecasting layer is used for solar energy variable prediction using XGBoost. The forecasting engine uses historical solar data and engineered features such as lag values, rolling averages, seasonal cycles and day-of-year features. XGBoost was selected because it performs well on structured time-series data and works well even when data is limited. This is a practical choice for software forecasting systems because it is fast to train, easy to update, interpretable, and computationally efficient. 

\qquad To obtain a forecast of a desired solar energy parameter (irradiance, temperature, or wind speed), the user selects the target variable and the number of days to forecast. Solar Intelligence dispatches the request to the XGBoost forecaster, which trains on the full NASA POWER historical record for the selected variable using the 10 engineered features described in Table 3 (calendar, cyclic seasonality, autoregressive lags at 1/7/30/365 days, and 7- and 30-day rolling means). The forecaster then iteratively rolls predictions forward: the model predicts day t + 1 using the observed lag values, appends this prediction to the historical series, and uses the augmented series to predict day t + 2, and so on until the requested horizon is reached. The dashboard returns the last observed value from the training data alongside a table of predicted values with their corresponding dates, and renders these as a time-series chart. This forecasting module operates independently of the RAG pipeline and is exposed through both the Streamlit Forecast tab and the /forecast REST endpoint.

\subsection{Interface and API Layer}

\qquad The interface layer provides access to the system. FastAPI is used for backend APIs. It handles request routing, API responses, endpoint management and service integration. Streamlit is used for the user-facing dashboard. The dashboard provides a natural language query interface, data comparison views, forecast visualization and evidence panels. This layer makes the system usable for both technical users and software integration.

\begin{table}[H]
\centering
\caption{System component and technology stack of Solar Intelligence}
\label{tab:stack}
\begin{tabular}{ll}
\toprule
\textbf{Component} & \textbf{Technology} \\
\midrule
Backend & FastAPI \\
Storage & DuckDB \\
Sparse retrieval & BM25 \\
Dense retrieval & BGE-small \\
Fusion & Reciprocal Rank Fusion \\
Chunking & PyMuPDF \\
Generation & llama3.2:3b via Ollama \\
Forecasting & XGBoost \\
Frontend & Streamlit \\
\bottomrule
\end{tabular}
\end{table}

\subsection{System Workflow}

\qquad This section explains how the system operates from beginning to end. First the user enters a natural-language query. The query is interpreted and classified, and the relevant system component is selected: structured data, scientific documents, or hybrid. They can select a solar energy variable to forecast. Results are processed, evidence and citations are incorporated, the response is generated, and the result is presented through the Streamlit dashboard.

\subsection{Deployment Architecture}

\qquad The complete system was deployed on Jetstream2 across two servers: a FastAPI server (with 13 REST endpoints, OpenAPI/Swagger documentation, and X-API-Key header authentication) and a Streamlit dashboard  (with five tabs: Ask, Compare, Forecast, Data Coverage, and About). A daily cron job on HPC host executes the incremental NASA POWER loader, keeping the dataset current within satellite-processing lag. The system exposes an MCP-style (Model Context Protocol) manifest advertising nine callable tools, their input schemas, authentication requirements, and example invocations. This manifest allows other AI agents to discover and call Solar Intelligence programmatically without prior coordination, supporting the broader aim of inter-agent collaboration. 
\section{Results and Discussion}

\subsection{Query Types and QA Citation Results}

\qquad As discussed earlier in the query layer the user can ask three types of queries: (1) semantic query (``How does dust affect solar panel efficiency?''), (2) structured query (``What was the average GHI in 2025?'') and (3) hybrid query (``Compare NASA vs Biosphere 2 irradiance for May 2013 and explain the difference''). Solar Intelligence outputs for semantic and structured queries are given below. 

\subsubsection{Semantic Query}

\begin{querybox}[]
\textbf{Query 1:} What is photovoltaic efficiency?\\[2pt]

\textbf{Answer}\\
Photovoltaic (PV) systems convert sunlight directly into electricity using
semiconductor cells. Efficiency of commercial panels ranges from 15\% to 22\%. [1]
According to [2], the thermally induced blue-shifted photoluminescence of a
low-bandgap absorber is coupled to a high-bandgap photovoltaic cell, resulting
in maximal theoretical conversion efficiencies of up to 69\%. Additionally,
research has demonstrated tenfold thermal-enhancement of useful radiation for
high-bandgap cells and 107\% enhancement in average photon energy. [2]
.\\[2pt]

\textbf{Type:} \texttt{semantic} \hspace {1em}
\textbf{Source:} \texttt{n/a} \hspace{1em}
\textbf{Sources:} 5\\[2pt]

\textbf{Cited Evidence}\\

{[1]} Project doc --- \emph{PV Systems: ``Photovoltaic (PV) systems convert
sunlight directly into electricity using semiconductor cells. Efficiency of
commercial panels ranges from 15\% to 22\%. Temperature negatively affects PV
output; for every degree Celsius above 25$^\circ$C, efficiency drops by about
0.4\%. Proper tilt and orientation significa\ldots''}\\

{[2]} Research paper --- \emph{Optical Refrigeration for Ultra-Efficient
Photovoltaics} (\url{https://arxiv.org/abs/1404.7345v1})
\end{querybox}

\begin{quote}

\qquad The Query 1 (``What is photovoltaic efficiency?'') is semantic in nature, and the classifier routed it accordingly rather than to the SQL path. Retrieval drew on both a project document and a research paper, and the generated figure for commercial panel efficiency (15--22\%) matches the retrieved source exactly, with no fabricated values. The response nonetheless illustrates two limitations of the semantic path. First, it omitted an operationally relevant fact present in the retrieved evidence, namely that efficiency falls by approximately 0.4\% for every degree Celsius above 25\,$^{\circ}$C. Second, it foregrounded a theoretical conversion ceiling of 69\% from an optical-refrigeration study without distinguishing that speculative bound from the measured commercial range, which risks misleading a reader who asked about present-day efficiency. 

\qquad This demonstrates the intended end-
to-end behavior of the semantic path (correct routing, cross-document retrieval,
and grounded generation without hallucination), while also illustrating a known
limitation: with a small locally hosted 3B-parameter model, generation prioritizes
distinctive numerical claims over comprehensive coverage of the retrieved evidence
and does not always surface the most operationally relevant facts. Improving
prompt engineering to guide the model toward practical relevance, or replacing
the generator with a larger model, is scoped for future work. 
\subsubsection{Structured Query}
\begin{querybox}[]
\textbf{Query 2:} Average GHI in March 2025\\[2pt]

\textbf{Answer} \\
The average irradiance from NASA was \textbf{5.95 kWh\,m$^{-2}$\,day$^{-1}$}\\[2pt] for the period 2025-03-01 to 2025-03-31.\\ 

For context, see [1] Solar Basics.\\

\textbf{Type:} \texttt{structured} \hspace{1em}
\textbf{Source:} \texttt{nasa} \hspace{1em}
\textbf{Sources:} 5\\[2pt]

\textbf{Structured data} \\

Source: \texttt{nasa} $|$
Aggregation: \texttt{avg} $|$
Value: 5.95 kWh\,m$^{-2}$\,day$^{-1}$\\[2pt]

\textbf{Cited Evidence} \\

[1] Project doc --- \emph{Solar Basics}
\end{querybox}

\begin{querybox}[]
\textbf{Query 3:} Average wind speed at Biosphere 2 in March 2013\\[2pt]\\
\textbf{Answer} \\
The average windspeed from B2 was \textbf{1.258 m\,s$^{-1}$}for the period
2013-03-01 to 2013-03-31. \\

For context, see [1] Nasa Vs B2 Validation.\\

\textbf{Type:} \texttt{structured} \hspace{1em}
\textbf{Source:} \texttt{b2} \hspace{1em}
\textbf{Sources:} 5\\[2pt]\\
\textbf{Structured data} \\
Source: \texttt{b2} $|$
Aggregation: \texttt{avg} $|$
Value: 1.258 m\,s$^{-1}$ $|$
Period: 2013-03-01 to 2013-03-31\\[2pt]\\
\textbf{Cited evidence:} [1] Project doc --- \emph{NASA vs.\ B2 Validation}
\end{querybox}

\qquad Query 2 (``Average GHI in March 2025'') demonstrates the system's structured path for satellite-derived NASA POWER data. The classifier assigned \texttt{type = structured}, \texttt{source = nasa}, with intent parameters \texttt{metric = irradiance} (the GHI field in the solar table), \texttt{aggregation = average}, \texttt{range = 2025-03}. The pipeline dispatched the query to the SQL engine, which executed a DuckDB \texttt{AVG()} over the daily GHI values for that range and returned 5.95\,kWh\,m$^{-2}$\,day$^{-1}$. The Solar Basics project note was surfaced as contextual evidence. As with all structured queries, the numerical answer was constructed directly from the SQL result via a deterministic template rather than composed by the LLM, so the reported value cannot be fabricated. 

For Query 3 (``Average wind speed at Biosphere 2 in March 2013'') the classifier assigned \texttt{type = structured}, \texttt{source = b2}, with intent parameters \texttt{metric = wind\_speed}, \texttt{aggregation = average}, \texttt{range = 2013-03-01} to \texttt{2013-03-31}. The pipeline dispatched the query to the SQL engine, which executed a DuckDB \texttt{AVG()} over the \texttt{b2\_solar} table and returned 1.258\,m\,s$^{-1}$. The NASA vs.\ B2 validation project note was surfaced as contextual evidence. As in Query 2, the numerical answer bypassed the LLM and came directly from the SQL result, which removes the risk of a fabricated value at the cost of a response with limited narrative content.

\subsection{NASA POWER and Biosphere 2 Comparison}

\qquad The Biosphere 2 record provides an in-situ reference for validating the NASA POWER product at the study site, with 312 overlapping days used for the cross-source evaluation. Figure 1 compares monthly averages from NASA and Biosphere 2 for irradiance (left), air temperature (center), and wind speed (right). 
\begin{figure}[H]
\centering
\includegraphics[width=0.8\linewidth]{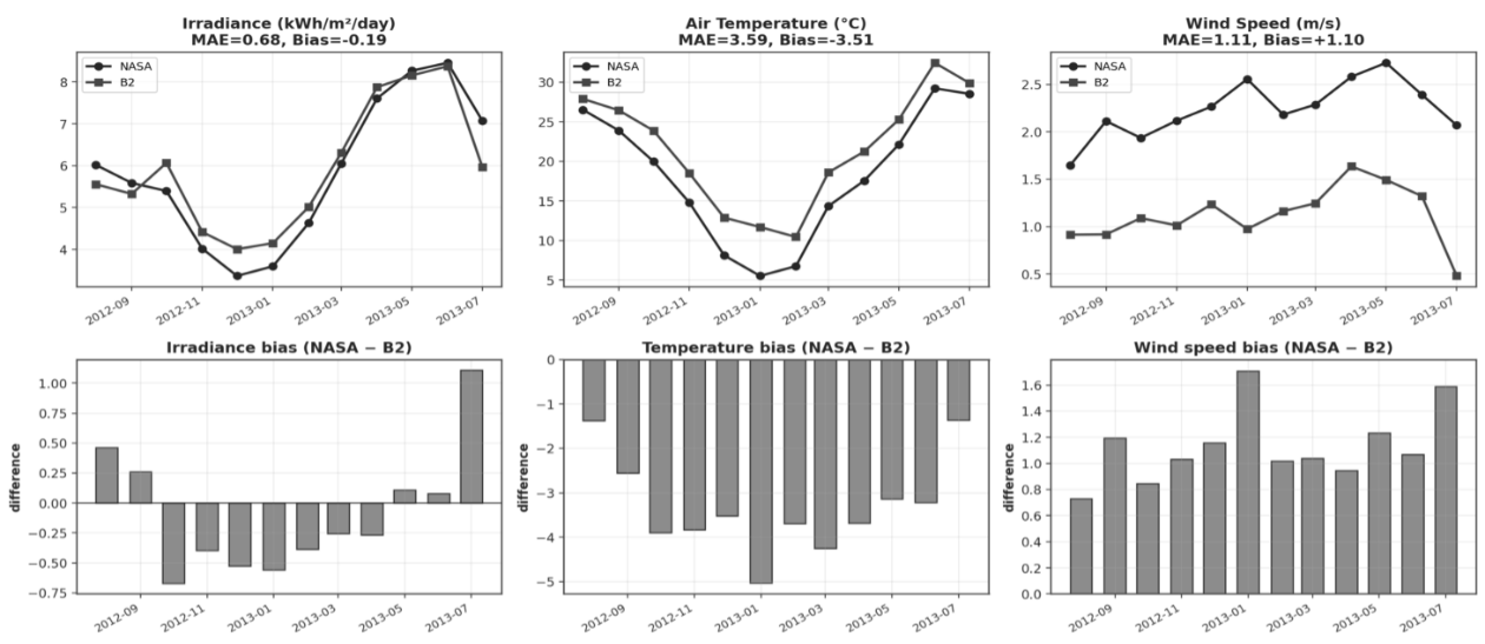}
\caption{ NASA POWER vs Biosphere 2 (B2) (32.578$^{\circ}$N, 110.848$^{\circ}$W) (Aug 2012 -- July 2013).}
\end{figure}
NASA captures the overall seasonal trend well in all three variables, but with different levels of accuracy. Irradiance shows the strongest agreement, with both sources following similar seasonal patterns and only small monthly differences, making it the most reliable variable. Temperature follows the same trend, but NASA is consistently lower than Biosphere 2, showing a clear cold bias across most months. Wind speed shows the weakest agreement, with NASA consistently overestimating compared to Biosphere 2 and showing larger month-to-month differences. Overall, NASA data is useful for capturing broad seasonal patterns, but local ground validation is necessary, especially for temperature and wind.
\subsection{Forecast Outputs}
\qquad

\begin{figure}[H]
\centering
\includegraphics[width=0.8\linewidth]{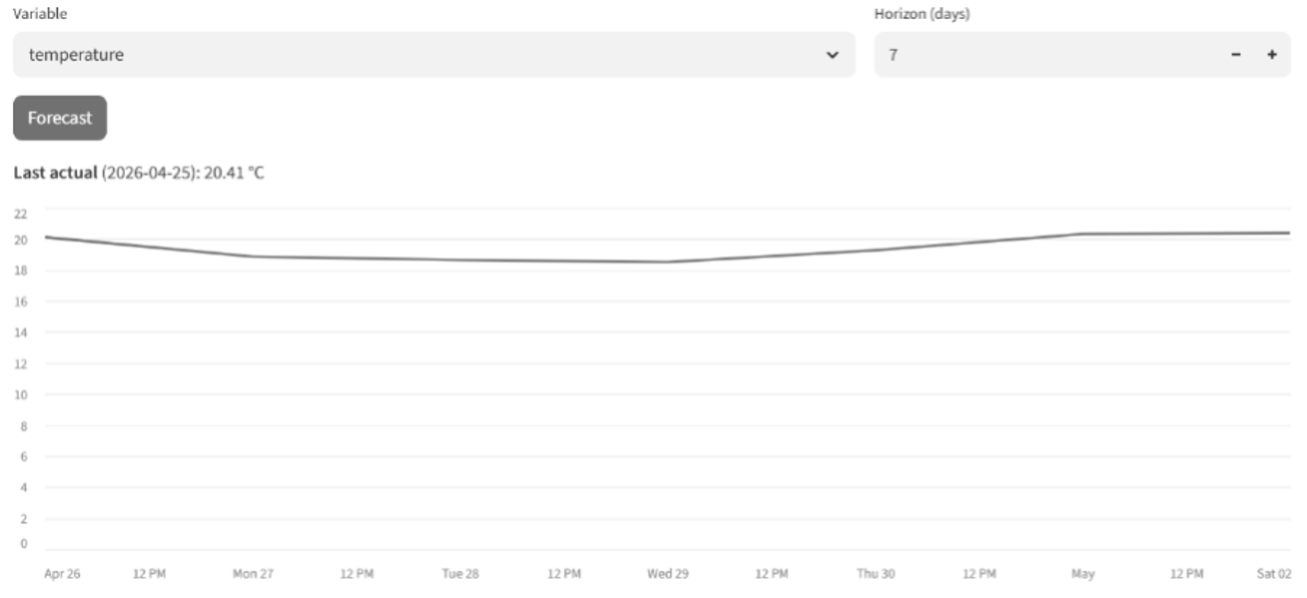}
\includegraphics[width=0.8\linewidth]{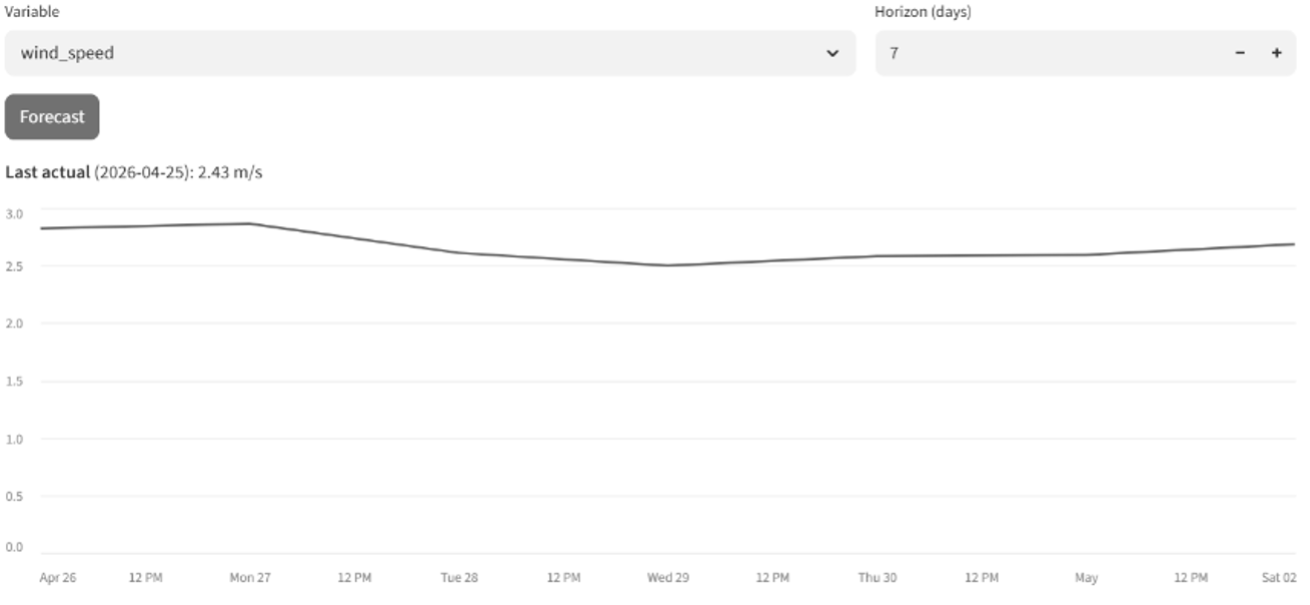}
\includegraphics[width=0.8\linewidth]{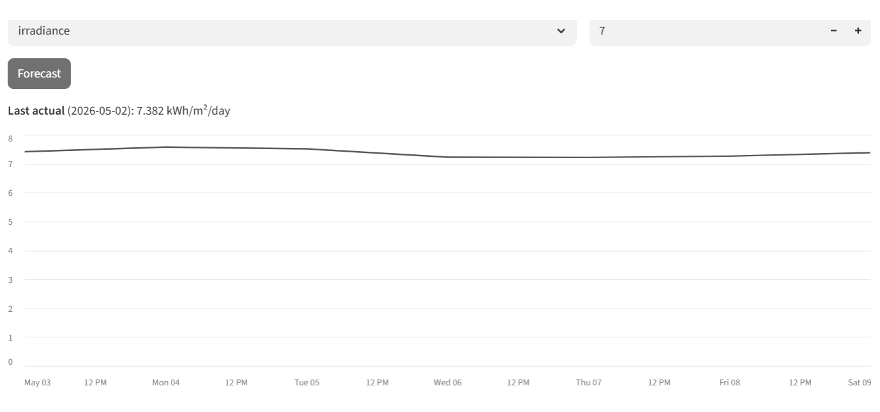 }
\caption{Seven-day forecasts from the Solar Intelligence Streamlit dashboard for temperature (top),  temperature (middle), and irradiance(bottom) at Biosphere 2 location. Each panel shows the last observed NASA POWER value at the top and the XGBoost roll-forward prediction as a continuous line (32.578$^{\circ}$N, 110.848$^{\circ}$W).}
\label{fig:validation}
\end{figure} 
 \qquad Figure 2 shows three 7-day forecasts generated by the Solar Intelligence dashboard for Biosphere 2 location: air temperature (top), wind speed (middle), and irradiance (bottom). Each forecast was produced by selecting the target variable and horizon length in the Forecast tab, at which point the XGBoost forecaster was trained on the full NASA POWER historical record for the selected variable and rolled predictions forward day-by-day. The wind speed forecast starting from a last observed value of 2.43 m/s on 25 April 2026 predicts a range of approximately 2.5–2.9 m/s across the following week with a mid-week dip. The irradiance forecast from a last observed value of 7.382 kWh/m²/day on 2 May 2026 predicts a near-flat trajectory around 7.4–7.6 kWh/m²/day. The temperature forecast from 20.41 °C on 25 April 2026 predicts a mild cooling early in the week to around 18.5 °C followed by warming back to ~20.5 °C by the weekend. Each panel also displays the last observed value at the top of the chart. The forecasts for all three variables appear to be in a valid range - no negative irradiance, no impossible wind speeds - but detailed interpretation is required to understand these forecasts, and evaluation against real values is also required before relying on the system outputs.

\section{Limitations}

\qquad The current Solar Intelligence system has several limitations. First, the location is hardcoded for Biosphere 2, so the system forecasts for that single site only. Second, NASA POWER data is not available live - there is a lag of about three days - so forecasts are made from the older data and it can introduce errors. Third, currently the system produces point forecasts rather than uncertainty forecasts, which limits its usefulness for reliable grid operations.

\section{Conclusion and Future Work}

\qquad Solar Intelligence shows that evidence-based retrieval over a research corpus, data can be combined with ML forecasting (XGBoost) and deployed as a single system. It was implemented with a FastAPI backend and a Streamlit dashboard and uses NASA POWER and Biosphere 2 ground sensor data. The Q/A layer returns satisfactory evidence-based answers for semantic, structured, and hybrid queries. The system currently produces short-, medium-, and long-range forecasts based on the user's selection. No evaluation of the forecasts has been performed; this can be done in the future versions. The system can benefit solar energy users and researchers by bringing fragmented resources into one place, saving time and providing insights for solar energy decision-making. The next first step is to enlarge the corpus for better Q/A. Second, the hardcoded location should be removed and the system generalized to multiple sites so that it can be tested and used for different locations. Third, there is scope to improve forecast quality and to adopt a probabilistic  and Bayesian approach (Singh, 2026) to make the system uncertainty-aware and the output usable for grid operations.

\section{Release}

\qquad The more information about the Solar Intelligence system can be found in the given repository: \url{https://github.com/jyotsnasingh11217/Solar-Intelligence} 

\section{Acknowledgements}

\qquad The author thanks the University of Arizona for supporting this work, and gratefully acknowledges Ash Black (College of Information Science) and Jeff Larsen (Biosphere 2) for their support. Computational resources were provided by the Jetstream2 HPC.


\section{References}

\qquad Appelbaum, Joseph, and Assaf Peled. ``Anisotropic diffuse radiation model of photovoltaic systems.'' \textit{Progress in Photovoltaics: Research and Applications} (2026).

\qquad Atsbeha, Amaha Kidanu, et al.\ ``Assessment of solar resource potential and estimation of direct and diffuse solar irradiation from sunshine hours data for Eastern Zone of Tigray, Northern Ethiopia.'' \textit{Sustainable Energy Research} 12.1 (2025): 16.

\qquad Chen, Tianqi, and Carlos Guestrin. ``XGBoost: A scalable tree boosting system.'' \textit{Proceedings of the 22nd ACM SIGKDD International Conference on Knowledge Discovery and Data Mining} (2016): 785--794.

\qquad Cormack, Gordon V., Charles L.A. Clarke, and Stefan Buettcher. ``Reciprocal rank fusion outperforms condorcet and individual rank learning methods.'' \textit{Proceedings of the 32nd International ACM SIGIR Conference on Research and Development in Information Retrieval}. 2009.

\qquad Fohagui, Fodoup Cyrille Vincelas, et al.\ ``Prediction of Cameroon's global solar radiation using deep learning and machine learning algorithms.'' \textit{Journal of Atmospheric and Solar-Terrestrial Physics} (2026): 106733.

\qquad Gandhi, Oktoviano, et al.\ ``The value of solar forecasts and the cost of their errors: a review.'' \textit{Renewable and Sustainable Energy Reviews} 189 (2024): 113915.

\qquad Gao, Yunfan, et al.\ ``Retrieval-augmented generation for large language models: a survey.'' \textit{arXiv preprint arXiv:2312.10997} (2023).

\qquad Grinsztajn, L\'{e}o, Edouard Oyallon, and Ga\"{e}l Varoquaux. ``Why do tree-based models still outperform deep learning on typical tabular data?'' \textit{Advances in Neural Information Processing Systems} 35 (2022): 507--520

\qquad Lewis, Patrick, et al.\ ``Retrieval-augmented generation for knowledge-intensive NLP tasks.'' \textit{Advances in Neural Information Processing Systems} 33 (2020): 9459--9474.

\qquad Mayer, Martin J\'anos, Dazhi Yang, and D\'avid Markovics. ``Quality and economic value of operational photovoltaic power forecasts in the day-ahead market.'' \textit{Energy Conversion and Management} 358 (2026): 121451.

\qquad Ouyang, Jing, et al.\ ``Seasonal distribution analysis and short-term PV power prediction method based on decomposition optimization Deep-Autoformer.'' \textit{Renewable Energy} 246 (2025): 122903.

\qquad Robertson, Stephen E., and Steve Walker. ``Some simple effective approximations to the 2-Poisson model for probabilistic weighted retrieval.'' \textit{SIGIR'94: Proceedings of the Seventeenth Annual International ACM-SIGIR Conference on Research and Development in Information Retrieval, organised by Dublin City University}. London: Springer London, 1994.

\qquad Robertson, Stephen E., et al.\ ``Okapi at TREC-3.'' \textit{TREC}. National Institute of Standards and Technology (NIST), 1994.

\qquad Segado-Moreno, Leandro C., et al.\ ``Past, current and future solar radiation trends in Europe: multi-source assessment of the role of clouds and aerosols.'' \textit{Remote Sensing of Environment} 333 (2026): 115122.

\qquad Singh, Jyotsna, et al.\ ``Modelling monthly diffuse solar radiation fraction and its validity over the Indian sub-tropics.'' \textit{International Journal of Climatology} 33.1 (2013): 77--86.

\qquad Singh, Jyotsna, and Manoj Kumar. ``Solar radiation over four cities of India: trend analysis using Mann-Kendall test.'' \textit{International Journal of Renewable Energy Resources} 6 (2016): 1385--1395.

\qquad Singh, Jyotsna, et al.\ ``Analysis of aggregated solar PV output in South Africa.'' \textit{Journal of Clean Energy Technologies} 5.5 (2017): 378--382.

\qquad Singh, Jyotsna. \textit{Hierarchical Bayesian Modeling of Solar Irradiance under Extreme Weather in the Tucson Electric Power Region}. No.\ EGU26-22088. Copernicus Meetings, 2026.

\qquad Voyant, Cyril, et al.\ ``Time series modeling and large scale global solar radiation forecasting from geostationary satellites data.'' \textit{Solar Energy} 102 (2014): 131--142.

\qquad Wild, Martin, et al.\ ``Global dimming and brightening: an update beyond 2000.'' \textit{Journal of Geophysical Research: Atmospheres} 114.D10 (2009).

\qquad Xiao, Shitao, et al.\ ``C-Pack: packed resources for general Chinese embeddings.'' \textit{Proceedings of the 47th International ACM SIGIR Conference on Research and Development in Information Retrieval}. 2024.
\end{quote}

\end{document}